\documentclass[11pt,a4paper]{article}
\usepackage[hyperref]{acl2019}
\usepackage{times}
\usepackage{latexsym}
\usepackage{amsmath}
\usepackage{hhline}
\usepackage{amssymb}
\usepackage{xcolor}
\usepackage{multirow}

\usepackage[english]{babel}
\usepackage[autostyle]{csquotes}
\usepackage{url}
\usepackage{graphicx}

\aclfinalcopy 

\title{Improving Visual Question Answering by Referring to \\ Generated Paragraph Captions}

\author{Hyounghun Kim \;\;\;\;\;\;\;\;\;\;\;\;\;\; Mohit Bansal \\
  UNC Chapel Hill \\
  {\tt \{hyounghk, mbansal\}@cs.unc.edu}}

\date{}

\begin{document}
\maketitle
\begin{abstract}
Paragraph-style image captions describe diverse aspects of an image as opposed to the more common single-sentence captions that only provide an abstract description of the image. These paragraph captions can hence contain substantial information of the image for tasks such as visual question answering. Moreover, this textual information is complementary with visual information present in the image because it can discuss both more abstract concepts and more explicit, intermediate symbolic information about objects, events, and scenes that can directly be matched with the textual question and copied into the textual answer (i.e., via easier modality match). Hence, we propose a combined Visual and Textual Question Answering (VTQA) model which takes as input a paragraph caption as well as the corresponding image, and answers the given question based on both inputs. In our model, the inputs are fused to extract related information by cross-attention (early fusion), then fused again in the form of consensus (late fusion), and finally expected answers are given an extra score to enhance the chance of selection (later fusion). Empirical results show that paragraph captions, even when automatically generated (via an RL-based encoder-decoder model), help correctly answer more visual questions. Overall, our joint model, when trained on the Visual Genome dataset, significantly improves the VQA performance over a strong baseline model.
\end{abstract}

\section{Introduction}
Understanding visual information along with natural language have been studied in different ways. In visual question answering (VQA) \cite{VQA, balanced_vqa_v2, lu2016hierarchical, fukui2016multimodal, xu2016ask, yang2016stacked, zhu2016visual7w, Anderson2017up-down}, models are trained to choose the correct answer given a question about an image. 
On the other hand, in image captioning tasks \cite{karpathy2015deep, densecap, Anderson2017up-down, krause2017hierarchical, liang2017recurrent, melaskyriazi2018paragraph}, the goal is to generate sentences which should describe a given image. Similar to the VQA task, image captioning models should also learn the relationship between partial areas in an image and the generated words or phrases. While these two tasks seem to have different directions, they have the same purpose: understanding visual information with language. If their goal is similar, can the tasks help each other?

In this work, we propose an approach to improve a VQA model by exploiting textual information from a paragraph captioning model. Suppose you are assembling furniture by looking at a visual manual. If you are stuck at a certain step and you are given a textual manual which more explicitly describes the names and shapes of the related parts, you could complete that step by reading this additional material and also by comparing it to the visual counterpart. With a similar intuition, paragraph-style descriptive captions can more explicitly (via intermediate symbolic representations) explain what objects are in the image and their relationships, and hence VQA questions can be answered more easily by matching the textual information with the questions. 

We provide a VQA model with such additional `textual manual' information to enhance its ability to answer questions. We use descriptive captions generated from a paragraph captioning model  which capture more detailed aspects of an image than a single-sentence caption (which only conveys the most obvious or salient single piece of information). 
We also extract properties of objects, i.e., names and attributes from images to create simple sentences in the form of ``[object name] is [attribute]". 
Our VTQA model takes these paragraph captions and attribute sentences as input in addition to the standard input image features. The VTQA model combines the information from text and image with early fusion, late fusion, and later fusion. With early fusion, visual and textual features are combined via cross-attention to extract related information. Late fusion collects the scores of candidate answers from each module to come to an agreement. In later fusion, expected answers are given an extra score if they are in the recommendation list which is created with properties of detected objects. Empirically, each fusion technique provides complementary gains from paragraph caption information to improve VQA model performance, overall achieving significant improvements over a strong baseline VQA model. We also present several ablation studies and attention visualizations.

\section{Related Work}
\noindent\textbf{Visual Question Answering (VQA)}:
VQA has been one of the most active areas among efforts to connect language and vision~\cite{malinowski2014multi, tu2014joint}. The recent success of deep neural networks, attention modules, and object plus salient region detection has made more effective approaches possible \cite{VQA, balanced_vqa_v2, lu2016hierarchical, fukui2016multimodal, xu2016ask, yang2016stacked, zhu2016visual7w,Anderson2017up-down}. 

\noindent\textbf{Paragraph Image Captioning}:
Another thread of research which deals with combined visual and language problem is the translation of visual contents to natural language. The first approach to this included using a single-sentence image captioning model \cite{karpathy2015deep}. However, this task is not able to accommodate the variety of aspects of a single image. \citet{densecap} expanded single-sentence captioning to describe each object in an image via a dense captioning model. Recently, paragraph captioning models \cite{krause2017hierarchical, liang2017recurrent,melaskyriazi2018paragraph} attempt to capture the many aspects in an image more coherently.

\begin{figure*}[ht]
  \includegraphics[width=\textwidth]{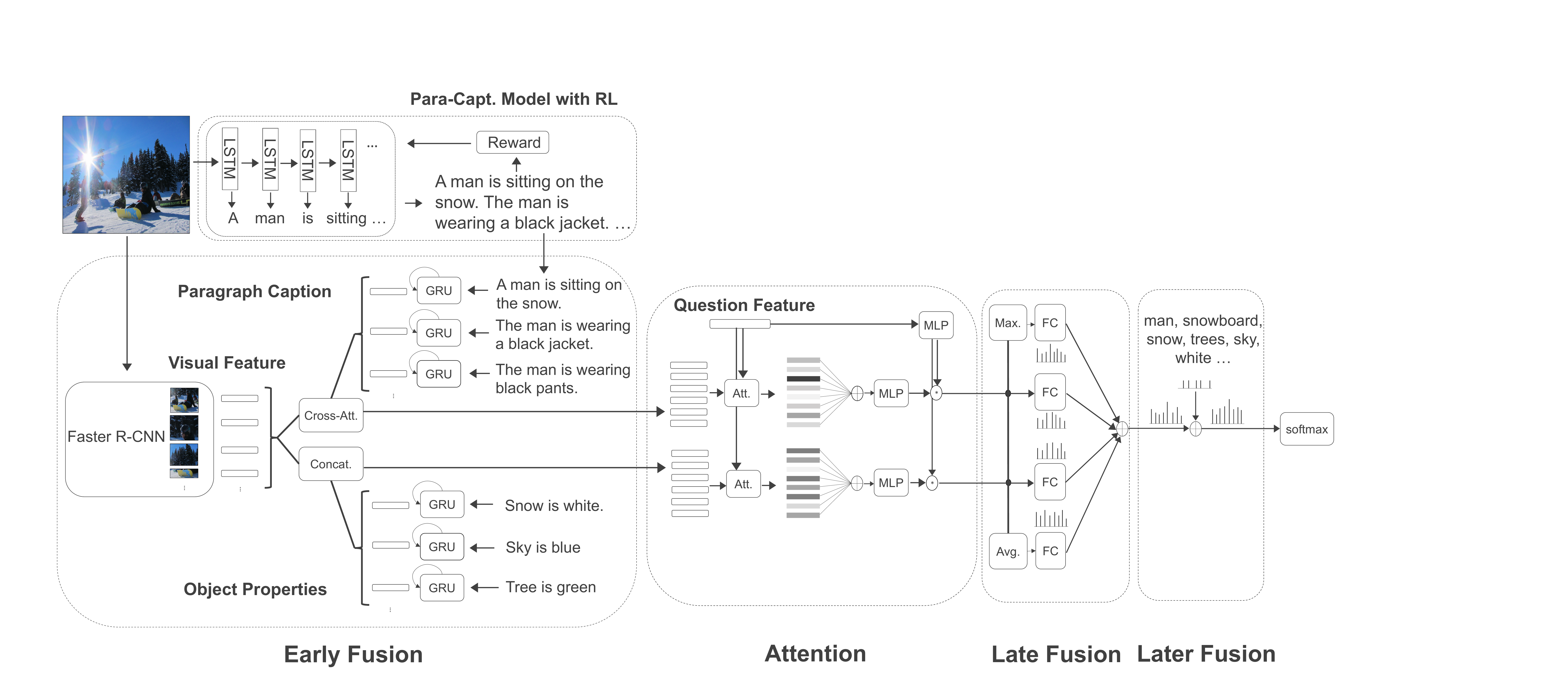}
  \vspace{-15pt}
\caption{VTQA Architecture: Early, Late, and Later Fusion between the Vision and Paragraph Features. \label{fig:model}}
\vspace{-10pt}
\end{figure*}

\section{Models} 
\vspace{-5pt}
The basic idea of our approach is to provide the VQA model with extra text information from paragraph captions and object properties (see Fig.~\ref{fig:model}).  

\subsection{Paragraph Captioning Model}
Our paragraph captioning module is based on \citet{melaskyriazi2018paragraph}'s work, which uses CIDEr~\cite{vedantam2015cider} directly as a reward to train their model. They make the approach possible by employing self-critical sequence training (SCST)~\cite{rennie2017self}. However, only employing RL training causes repeated sentences. As a solution, they apply \(n\)-gram repetition penalty to prevent the model from generating such duplicated sentences. We adopt their model and approach to generate paragraph captions.

\subsection{VTQA Model}

\subsubsection{Features}
\noindent\textbf{Visual Features}: We adopt the bottom-up and top-down VQA model from \newcite{Anderson2017up-down}, which uses visual features from the salient areas in an image (bottom-up) and gives them weights using attention mechanism (top-down) with features from question encoding. Following \citet{Anderson2017up-down}, we also use Faster R-CNN \cite{ren2015faster} to get visual features \(V \in \mathbb{R} ^{O\times d}\), where \(O\) is \#objects detected and \(d\) is the dimension of each visual feature of the objects.

\noindent\textbf{Paragraph Captions}: These provide diverse aspects of an image by describing the whole scene.
We use GloVe \cite{pennington2014glove} for the word embeddings. The embedded words are sequentially fed into the encoder, for which we use GRU \cite{cho2014learning}, to create a sentence representation, \(s_{i} \in \mathbb{R}^d\): $s_i = \textrm{ENC}_{sent}(w_{0:T})$, where \(T\) is the number of words. The paragraph feature is a matrix which contains each sentence representation in each row, \(P \in \mathbb{R}^{K\times d} \), where \(K\) is the number of sentences in a paragraph. 

\noindent\textbf{Object Property Sentences}: The other text we use is from properties of detected objects in images (name and attribute), which can provide explicit information of the corresponding object to a VQA model. We create simple sentences like, ``[object name] is [attributes]". We then obtain sentence representations by following the same process as what we do with the paragraph captions above.
Each sentence vector is then attached to the corresponding visual feature, like \lq name tag\rq{}, to allow the model to identify objects in the image and their corresponding traits.

\subsubsection{Three Fusion Levels}
\noindent\textbf{Early Fusion}: In the early fusion stage, visual features are fused with paragraph caption and object property features to extract relevant information. For visual and paragraph caption features, cross-attention is applied to get similarity between each component of visual features (objects) and a paragraph caption (sentences). We follow \citet{seo2016bidirectional}'s approach to compute the similarity matrix, \(S \in \mathbb{R}^{O\times K} \). From the similarity matrix $V^p = \textrm{softmax}(S^{T})V$ and the new paragraph representation, \(P^f\) is obtained by concatenating \(P\) and \(P*V^p\): $P^f = [P;P*V^p]$, where * is element-wise product operation. For visual feature and object property feature \(C\), they are already aligned and the new visual feature \(V^f\) becomes $V^f = [V;V*C]$. Given the fused representations, the attention mechanism is applied over each row of the representations to weight more relevant features to the question.
\vspace{-5pt}
\begin{align}
    a_i &= w_a^T(\textrm{ReLU}(W_{sa}s^f_i) * \textrm{ReLU}(W_{qa}q)) \\
    \mathbf{\alpha} &= \textrm{softmax}(a)
\end{align}
where, \(s^f_i\) is a row vector of new fused paragraph representation and \(q\) is the representation vector of a question which is encoded with GRU unit. \(w_a^T\), \(W_{sa}\), and \(W_{qa}\) are trainable weights. Given the attention weights, the weighted sum of each row vector, \(s^f_i\) leads to a final paragraph vector $p = \sum_{i=1}^K \alpha_is^f_i$. The paragraph vector is fed to a nonlinear layer and combined with question vector by element-wise product.
\vspace{-5pt}
\begin{align}
    p^q &= \textrm{ReLU}(W_{p}p) * \textrm{ReLU}(W_{q}q) \\
    L_p &= \textrm{classifier}(p^q)
\end{align}
where \(W_{p}\) and \(W_{q}\) are trainable weights, and \(L_p\) contains the scores for each candidate answer.
The same process is applied to the visual features to obtain $L_v = \textrm{classifier}(v^q)$.

\noindent\textbf{Late Fusion}: In late fusion, logits from each module are integrated into one vector. We adopt the approach of~\newcite{wang2016temporal}. 
Instead of just adding the logits, we create two more vectors by max pooling and averaging those logits and add them to create a new logit  $L_{new} = L_1 + L_2 + ... + L_n + ... + L_{max} + L_{avg}$, where \(L_{n}\) is \(n\)th logit, and \(L_{max}\) and \(L_{avg}\) are from max-pooling and averaging all other logits. The intuition of creating these logits is that they can play as extra voters so that the model can be more robust and powerful.

\noindent\textbf{Answer Recommendation or `Later Fusion'}:  Salient regions of an image can draw people's attention and thus questions and answers are much more likely to be related to those areas. Objects often denote the most prominent locations of these salient areas. From this intuition, we introduce a way to directly connect the salient spots with candidate answers. We collect properties (name and attributes) of all detected objects and search over answers to figure out which answer can be extracted from the properties. Answers in this list of expected answers are given extra credit to enhance the chance to be selected. 
If logit \(L_{\text{before}}\) from the final layer contains scores of each answer, we want to raise the scores to logit \(L_{\text{after}}\) if the corresponding answers are in the list \(l_c\):
\begin{equation}
\begin{split}
&L_{\text{before}} = \{a_1, a_2, ..., a_n, ..\} \\
&L_{\text{after}} = \{\hat{a}_1, \hat{a}_2, ..., \hat{a}_n, ..\} \\
\end{split}
\end{equation}
\vspace{-15pt}
\begin{equation}
    \hat{a}_n = \left\{ \begin{array}{rl}
    a_n + c \cdot \textrm{std}(L_{before}) &\mbox{ if } n \in l_c\\
    a_n \qquad \qquad &\mbox{ otherwise}
    \end{array} \right.
\end{equation}
where the \(\textrm{std}(\cdot)\) operation calculates the standard deviation of a vector and \(c\) is a tunable parameter. \(l_c\) is the list of the word indices of detected objects and their corresponding attributes. The indices of the objects and the attributes are converted to the indices of candidate answers.

\section{Experimental Setup}

\noindent\textbf{Paragraph Caption}: We use paragraph annotations of images from Visual Genome \cite{krishnavisualgenome} collected by \citet{krause2017hierarchical}, since this dataset is the only dataset (to our knowledge) that annotates long-form paragraph image captions. We follow the dataset split of 14,575 / 2,487 / 2,489 (train / validation / test).

\noindent\textbf{Visual Question Answering Pairs}: We also use the VQA pairs dataset from Visual Genome so as to match it with the provided paragraph captions. We almost follow the same image dataset split as paragraph caption data, except that we do not include images that do not have their own question-answer pairs in the train and evaluation sets. The total number of candidate answers is 177,424. Because that number is too huge to train, we truncate the question-answer pairs whose answer's frequency are under 30, which give us a list of 3,453 answers. So, the final number of  question-answering pairs are 171,648 / 29,759 / 29,490 (train / validation / test).

\noindent\textbf{Training Details}: Our hyperparameters are selected using validation set. The size of the visual feature of each object is set to 2048 and the dimension of the hidden layer of question encoder and caption encoder are 1024 and 2048 respectively. We use AdaMax~\cite{kingma2014adam} for the optimizer and a learning rate of 0.002. We modulate the final credit, which is added to the final logit of the model, by multiplying a scalar value \(c\) (we tune this to 1.0).

\section{Results, Ablations, and Analysis}
\vspace{-5pt}

\begin{table}[t]
\small
\begin{center}

 \begin{tabular}{|c|l|c|}
  \hline
 & Model & Test accuracy (\%)\\
 \hline
 1 & VQA baseline & 44.68 \\
 2 & VQA + MFB baseline & 44.94 \\
 \hline
 3 & VTQA (EF+LF+AR) & 46.86 \\
  \hline
\end{tabular}
\end{center}
\vspace{-10pt}
\caption{Our VTQA model significantly outperforms (\(p < \) 0.001) the strong baseline VQA model (we do not apply MFB to our VTQA model, since it does not work for the VTQA model). \label{tbl:VQAvsVTQA}} 
\vspace{-5pt}
\end{table}

\begin{table}[t]
\small
\begin{center}
 \begin{tabular}{|c|l|c|}
  \hline
 & Model &  Val accuracy (\%) \\
 \hline
    1 & VTQA + EF (base model) & 45.41\\
    2 &  VTQA + EF + LF & 46.36\\
    3 &  VTQA + EF + AR & 46.95\\
    4 & VTQA + EF + LF + AR & 47.60\\
  \hline
\end{tabular}
\end{center}
\vspace{-10pt}
\caption{Our early (EF), late (LF), and later fusion (or Answer Recommendation AR) modules each improves the performance of our VTQA model.\label{tbl:eflfar}} 
\vspace{-10pt}
\end{table}

\paragraph{VQA vs. VTQA}
As shown in Table \ref{tbl:VQAvsVTQA}, our VTQA model increases the accuracy by 1.92\% from the baseline VQA model for which we employ \citet{Anderson2017up-down}'s model and apply multi-modal factorized bilinear pooling (MFB) \cite{yu2017multiMFB}. This implies that our textual data helps improve VQA model performance by providing clues to answer questions. We run each model five times with different seeds and take the average value of them. For each of the five runs, our VTQA model performs significantly better (\(p<0.001\)) than the VQA baseline model.

\paragraph{Late Fusion and Later Fusion Ablations}
As shown in row 2 of Table \ref{tbl:eflfar}, late fusion improves the model by 0.95\%, indicating that visual and textual features complement each other. As shown in row 3 and 4 of Table \ref{tbl:eflfar}, giving an extra score to the expected answers increases the accuracy by 1.54\% from the base model (row 1) and by 1.24\% from the result of late fusion (row 2), respectively. This could imply that salient parts (in our case, objects) can give direct cues for answering questions.\footnote{Object Properties: Appending the encoded object properties to visual features improves the accuracy by 0.15\% (47.26 vs. 47.41). This implies that incorporating extra textual information into visual features could help a model better understand the visual features for performing the VQA task.}

\begin{table}[t]
\small
\begin{center}
 \begin{tabular}{|c|l|c|}
  \hline
 & Model &  Val accuracy (\%) \\
 \hline
    1 & TextQA with GT  & 43.96\\
    2 & TextQA with GenP & 42.07 \\
   \hline
\end{tabular}
\end{center}
\vspace{-10pt}
\caption{ TextQA with GT model outperforms TextQA with GenP (we run each model five times with different seeds and average the scores. GT: Ground-Truth, GenP: Generated Paragraph). \label{tbl:textQA}
} 
\vspace{-10pt}
\end{table}

\begin{table}[t]
\small
\begin{center}
 \begin{tabular}{|c|l|c|}
  \hline
  & Human Eval. &  Accuracy (\%) \\
  \hline
   1 & with GT & 55.00\\
   2 & with GenP & 42.67\\
   \hline
\end{tabular}
\end{center}
\vspace{-10pt}
\caption{ Human evaluation only with paragraph captions and questions of the validation dataset. Human evaluation with GT shows better performance than human evaluation with GenP. \label{tbl:humanEval}
} 
\vspace{-10pt}
\end{table}

\vspace{-2pt}
\paragraph{Ground-Truth vs. Generated Paragraphs \label{para:textqa}}
We manually investigate (300 examples) how many questions can be answered only from the ground-truth (GT) versus generated paragraph (GenP) captions. We also train a TextQA model (which uses cross-attention mechanism between question and caption) to evaluate the performance of the GT and GenP captions. As shown in Table \ref{tbl:textQA}, the GT captions can answer more questions correctly than GenP captions in TextQA model evaluation. Human evaluation with GT captions also shows better performance than with GenP captions as seen in Table \ref{tbl:humanEval}. However, the results from the manual investigation have around 12\% gap between GT and generated captions, while the gap between the results from the TextQA model is relatively small (1.89\%). This shows that paragraph captions can answer several VQA questions but our current model is not able to extract the extra information from the GT captions. This allows future work: (1) the TextQA/VTQA models should be improved to extract more information from the GT captions; (2) paragraph captioning models should also be improved to generate captions closer to the GT captions.\footnote{We also ran our full VTQA model with the ground truth (GT) paragraph captions and got an accuracy value of 48.04\% on the validation dataset (we ran the model five times with different seeds and average the scores), whereas the VTQA result from generated paragraph captions was 47.43\%. This again implies that our current VTQA model is not able to extract all the information enough from GT paragraph captions for answering questions, and hence improving the model to better capture clues from GT captions is useful future work.}

\begin{figure}[t]
  \includegraphics[width=80mm]{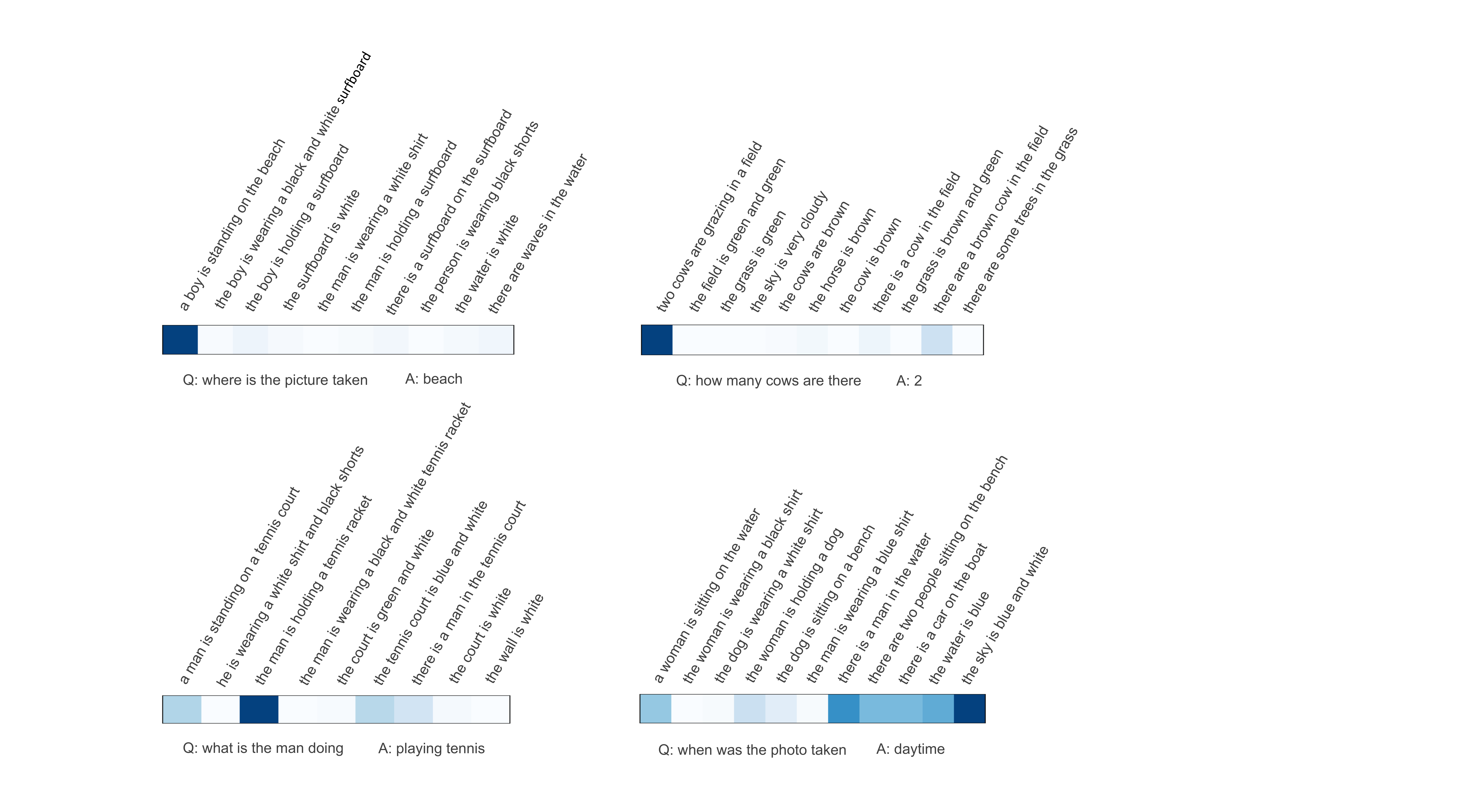}
  \vspace{-24pt}
\caption{Attention Visualization for an example answered correctly by our model. \label{fig:att_visu_1}}
\vspace{-15pt}
\end{figure}
\vspace{-5pt}
\paragraph{Attention Analysis}
Finally, we also visualize the attention over each sentence of an input paragraph caption w.r.t. a question. As shown in Figure~\ref{fig:att_visu_1}, a sentence which has a direct clue for a question get much higher weights than others. 
This explicit textual information helps a VQA model handle what might be hard to reason about only-visually, e.g., `two (2) cows'. Please see \autoref{app:visu} for more attention visualization examples.

\section{Conclusion}
We presented a VTQA model that combines visual and paragraph-captioning features to significantly improve visual question answering accuracy, via a model that performs early, late, and later fusion.
While our model showed promising results, it still used a pre-trained paragraph captioning model to obtain the textual symbolic information. In future work, we  are investigating whether the VTQA model can be jointly trained with the paragraph captioning model.

\section*{Acknowledgments}
We thank the reviewers for their helpful comments. This work was supported by NSF Award \#1840131, ARO-YIP Award \#W911NF-18-1-0336, and faculty awards from Google, Facebook, Bloomberg, and Salesforce. The views, opinions, and/or findings contained in this article are those of the authors and should not be interpreted as representing the official views or policies, either expressed or implied, of the funding agency.

\bibliography{acl2019}
\bibliographystyle{acl_natbib}

\appendix
\section*{Appendices}
\begin{figure*}[t]
  \includegraphics[width=\textwidth]{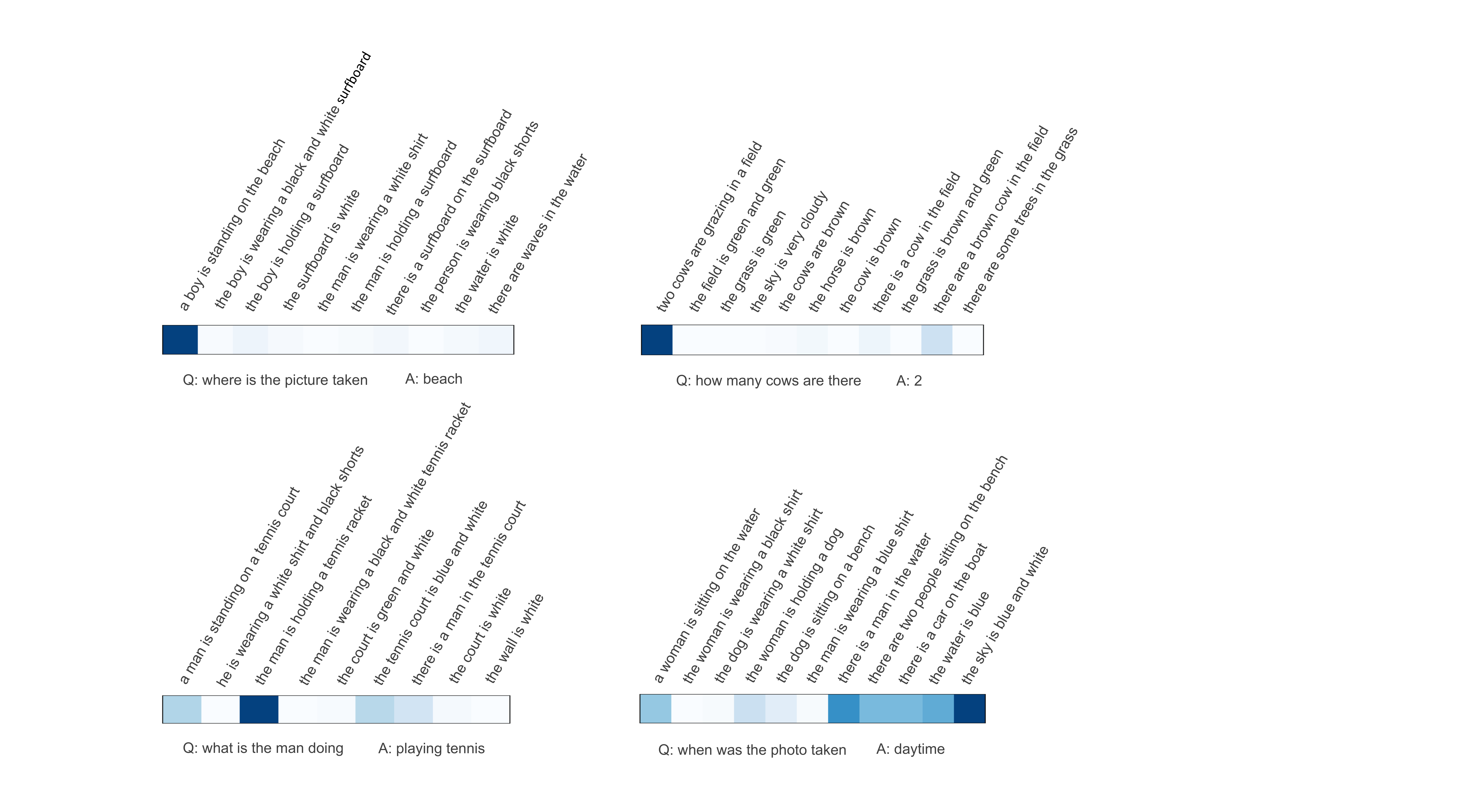}
\caption{Attention Visualization: For all examples, our model answers correctly. \label{fig:att_visu_sup}}
\end{figure*}

\section{Attention Visualization \label{app:visu}}
As shown in Figure~\ref{fig:att_visu_sup}, paragraph captions contain direct or indirect clues for answering questions. 
\paragraph{The upper left figure}
This is the case that a sentence in the paragraph caption can give obvious clue for answering the given question. By looking at the sentence ``a boy is standing on the beach'', this question can be answered correctly.
\paragraph{The upper right figure}
The sentence ``two cows are grazing in a field '' gives the correct answer ``2'' directly.
\paragraph{The bottom left figure}
There is no direct clue like ``he is playing tennis'', but the correct answer can be inferred by integrating the information from different sentences such as ``the man is holding a tennis racket'' and ``a man is standing on a tennis court''.
\paragraph{The bottom right figure}
This case seems tricky, but the answer can be inferred by associating the blue sky with daytime.

\end{document}